\documentclass{ifcolog-c}
\usepackage[utf8]{inputenc}
   
\numberwithin{definition}{section}

\usepackage{algorithm2e}
\usepackage{amsmath}
\usepackage{amsthm}
\usepackage{booktabs}
\usepackage{enumitem}
\usepackage{graphicx}
\usepackage{times}
\usepackage[nobiblatex]{xurl}
\usepackage{xcolor}
\usepackage{multirow}

\title{Argumentation and Machine Learning}
\titlerunning{Argumentation and Machine Learning}

\addauthor[a.rago@imperial.ac.uk]{Antonio Rago\footnote{Equal contribution.}}{Department of Computing, Imperial College London}
\addauthor[kristijonas.cyras@ericsson.com]{Kristijonas \v{C}yras$^*$}{Ericsson}
\addauthor[jack.mumford@liverpool.ac.uk]{Jack Mumford}{Department of Computer Science, University of Liverpool}
\addauthor[oana.cocarascu@kcl.ac.uk]{Oana Cocarascu}{Department of Informatics, King's College London}

\authorrunning{Rago, \v{C}yras, Mumford \& Cocarascu}

\titlethanks{}
\usepackage{fancyhdr}
\usepackage{natbib}
\setcitestyle{authoryear,open={(},close={)}}

\newcommand{\tags}[1]{{\textbf{#1}}}

\begin{document} 

\maketitle
\begin{abstract}

This chapter provides an overview of research works that present approaches with some degree of cross-fertilisation between Computational Argumentation and Machine Learning. Our review of the literature identified two broad themes representing the purpose of the interaction between these two areas: argumentation for machine learning and machine learning for argumentation. Across these two themes, we systematically evaluate the spectrum of works across various dimensions, including the type of learning and the form of argumentation framework used. Further, we identify three types of interaction between these two areas: synergistic approaches, where the Argumentation and Machine Learning components are tightly integrated; segmented approaches, where the two are interleaved such that the outputs of one are the inputs of the other; and approximated approaches, where one component shadows the other at a chosen level of detail. 
We draw conclusions about the suitability of certain forms of Argumentation for supporting certain types of Machine Learning, and vice versa, with clear patterns emerging from the review.
Whilst the reviewed works provide inspiration for successfully combining the two fields of research, we also identify and discuss limitations and challenges that ought to be addressed in order to ensure that they remain a fruitful pairing as AI advances.

\end{abstract}

\section{Introduction}

In this chapter, we overview research works that combine Computational Argumentation (henceforth simply argumentation) and Machine Learning (ML). 
In this section, we start with a general outlook of the themes and trends prevalent among the individual works overviewed in the chapter, including a loose categorisation of three types of interactions between ML an argumentation models, before briefly covering related work and papers we chose to omit.  
We then move on to our literature overview in §\ref{sec:overview} which we split by broad themes of the purposes of the interactions between ML and argumentation, briefly discussing promising research avenues.
Finally, we conclude in~§\ref{sec:conclusions}.

We assume the reader to be familiar with forms and frameworks of argumentation, e.g.\ as discussed in the 1\textsuperscript{st} Volume of the Handbook of Formal Argumentation \citep{handbook}, 
and fundamentals of and common methods in ML, as easily accessible e.g.\ on Wikipedia.

\subsection{Outlook}
\label{subsec:outlook}

Studies that consider interactions of argumentation and ML exhibit the following trends. 
Starting with ML characteristics, the \tags{type of learning} employed is largely \emph{supervised}, though there are instances of works that consider \emph{unsupervised} as well as both un/supervised types of learning, and a reasonable amount of works that focus on \emph{Reinforcement} Learning (RL). 
In terms of \tags{ML models} and algorithms employed, it is typical, but not exclusively so, to make use of simpler techniques and architectures, such as \emph{rule learning} (including both rule induction and rule extraction), 
\emph{tree-based models} (such as decision trees (DTs) and random forests (RFs)), 
\emph{naive Bayesian classifiers} (NBCs),
\emph{support vector machines} (SVMs), 
(usually shallow and/or feed-forward) \emph{neural networks} (NNs) and model-based \emph{RL}. 
Some recent works though opt for modern complex models, in particular \emph{graph neural networks} (GNNs), typically available off-the-shelf.
The \tags{data} that is used in experiments is most often of the \emph{tabular} type, though \emph{textual} and \emph{image} data also make their marks, while in the case of RL, \emph{simulations} in simple environments are prominent. 
The \tags{datasets} themselves are largely basic and small, often coming from the \emph{UCI ML Repository}.\footnote{\href{https://archive.ics.uci.edu}{https://archive.ics.uci.edu}} Though again, the recent works employing GNNs typically work on \emph{argument graph datasets} afforded by the International Competition on Computational Models of Argumentation (ICCMA)\footnote{\href{https://argumentationcompetition.org/}{https://argumentationcompetition.org/}}. 

In terms of \tags{argumentation frameworks}, we see a great diversity, ranging from the abstract to the structured kind. 
There is a notable focus on \emph{Abstract Argumentation Frameworks} (AFs), \emph{Value-Based Argumentation Frameworks} (VAFs) and (\emph{Quantitative}) \emph{Bipolar AFs} ((Q)BAFs), with QBAFs especially prominent in combination with NNs. 
Many works use \emph{Structured Argumentation} (SA), including the classical formalisms such as \emph{Defeasible Logic Programming}
(DeLP) and \emph{assumption-based argumentation} (ABA), but more prominently specific \emph{rule-based argumentation frameworks} (RB), often constructed using rule learning. 
Some \emph{Probabilistic Argumentation} (PA) and forms of \emph{informal argumentation} also appear. 
In terms of semantics, the most popular among the works overviewed seem to be those of the classic grounded or preferred extensions, in terms of both extensions and of sceptical and credulous acceptability, but gradual semantics are also often used, as are dialogues as both a form of argumentative interaction and inference. 
We will elaborate on such argumentative and ML-targeted characteristics of individual works in~§\ref{sec:overview}. 

Tangentially to the trends of characteristics of either ML or argumentation, we distinguish three \tags{types of interaction} between ML and argumentation. 
First, a \emph{synergistic} combination of an ML model and an argumentation model essentially means tightly integrating, or merging, the two kinds of models into one. 
The second, a \emph{segmented} approach, is where ML and argumentation are interleaved in performing learning and reasoning, using outputs of one as the inputs of another. 
Finally, the third type of interaction is \emph{approximated}, in the sense that one type of model is meant to approximate or shadow, at varying levels of detail, the other. 
We recognise that the boundaries between these notions are not clear cut and we do not aim at precise definitions of interactions. Rather, we posit these as guiding concepts, to glean the rough mode of interaction between argumentation and ML in each individual work, and hope that the individual descriptions given later in §\ref{sec:overview} will clarify these three notions. 

Two broad themes of the \tags{purpose of interaction} between argumentation and ML emerge. The first one that we identify is of using \emph{argumentation to improve and/or explain ML models}. We overview research works following this theme in §\ref{subsec:arg}. 
The other theme is that of using \emph{ML to support, analyse 
or replace 
argumentation}. We overview such themed works in §\ref{subsec:ml}. 

As for improving and/or explaining ML with argumentation, we identify the following motifs. 
SA is proposed to improve ML-based classification, often via rule-based/premise-conclusion kind of arguments constructed using rule learning, typically via segmented or synergistic approaches. 
This can be achieved by injecting knowledge into ML-based systems via rules or preferences, which is enabled by argumentation. 
Argumentation can also improve RL-based policies in Multi-Agent Systems, typically by means of argumentative selection of actions. 
Further, argumentation is often meant to explain ML-based systems, purportedly because of argumentative relations (especially in synergistic and segmented approaches) that are particularly analytical and interpretable.
This can be achieved for instance by reasoning argumentatively with rules learnt from data in a segmented or synergistic approach, or by argumentatively explaining why one classification decision is preferred over another after approximating an ML model. 
Explainability can also be enhanced with argumentation in RL by exhibiting argumentative interaction between agents' actions, or by using an argument graph as a structure to learn on and thereafter explain by means of existing argumentative explainability methods (see the recent surveys \citep{Cyras_21,Vassiliades_21,Guo_23}). 
It should be noted that some of the opportunities for improving ML models with argumentation which were foreseen in \citep{Longo_16} have indeed been realised in some cases, particularly those concerned with argumentative knowledge being injected into, and thus guiding inference in, ML models.

As for ML supporting argumentation, several main motifs emerge.
One is that of using ML to generate argumentation frameworks.  
This can be achieved by means of rule learning, particularly using Inductive ML, purely for constructing rule-based argumentation frameworks in a segmented fashion to reason with. (This echoes the theme of an interaction 
in the opposite direction, of argumentation for improving ML models, as mentioned above, even though the intrinsic goals of the research are distinct.)
Meanwhile, frameworks of the abstract kind, particularly AFs, can be learnt from other types of data or from other AFs themselves, often using approaches of the segmented kind. 
On the other hand, ML models can be used for the task of computing argumentation frameworks, meaning to predict the acceptance statuses of arguments, by and large in AFs using GNNs in an approximated fashion. 
Yet another recent motif is that of investigating if some forms of ML, particularly NNs, can be seen as argumentation frameworks, particularly QBAFs, and vice versa, for the purposes of advancing or even replacing such argumentation frameworks, as well as analysing or advancing such ML models. 
Finally, we encounter a couple of works that study argumentation in everyday life and investigate whether ML can help with recommending arguments to be put forward in real-world argumentative discussions, or whether ML should be used instead of formal argumentation due to its issues in modelling these situations. 

By the end of §\ref{sec:overview}, we will have seen many works where various forms of argumentation and ML models interact in different ways for one-way or mutual benefit. We will see different application areas (e.g.\  medical, robotics, prediction, legal and commonsense reasoning) and various tasks (e.g.\ classification, 
policy learning, argument graph generation and computation) addressed by the proposed systems. We will deem some research avenues to be more promising, often due to the use of modern ML and more extensive experimentation, others less so due to simplistic settings and/or empirical evaluation. 

A general criticism for argumentation in meeting ML is that of the apparent lack of implementations (in great contrast to the abundance of off-the-shelf ML models), which makes it harder for the interaction between the two to take place. The lack of user studies supporting the presumed benefits of argumentation is another crucial gap in the pairing between argumentation and ML. 
These concerns are implicit from our literature overview and we believe should be addressed if this cross-disciplinary research is to bear fruit.

Before delving into discussing individual works, we briefly mention related work, as well as omitted works that are out of scope of this chapter.

\subsection{Related Work and Omissions}
\label{subsec:related}

We will first discuss other reviews of the literature on argumentation which are closely linked to that which we provide.
The work with an objective which most closely aligns with our own is by \cite{Cocarascu_16}, in which the authors provide a comprehensive overview of argumentation's support for ML. The authors here note the relative youth of this area of research at the time, which indeed motivates the need for our contribution after many years of rapid progress in ML, both in general and combined with argumentation. 
One of the directions which we consider where there has indeed been a significant amount of progress is the use of argumentation for supporting explainable AI (concerning ML models or otherwise), which motivated three recent surveys of this literature by \cite{Vassiliades_21}, \cite{Cyras_21} and \cite{Guo_23}. 
It should be noted that cross-fertilisations between argumentation and either ML or explainable AI host a reasonable degree of overlap but neither is subsumed by the other.
Finally, \citeauthor{Proietti_23}~(\citeyear{Proietti_23}) give a roadmap towards Neuro-Argumentative Learning, i.e.\ learning argumentation frameworks from data, with a focus on neuro-symbolic approaches, looking ahead to directions for future work and challenges which remain.

With regards to works which were considered to be surplus to our review, the largest and most related such body of work was that on Argument Mining, and we refer the readers to the Chapter on Argument Mining for a comprehensive overview of these works.
We also leave out papers which consider tasks related to natural language processing or generation, e.g.\ works concerning the use of argumentation for analysis \citep{Hinton_23,Rajasekharan_23} or fine-tuning \citep{Thorburn_22,Furman_23} of Large Language Models \citep{Brown_20}.
Also, while argumentation frameworks can be seen as knowledge graphs of a specific form, we consider the wide range of cross-fertilisations between ML and knowledge graphs (e.g.\ see \citep{Tiddi_22} for a recent survey) to be outside the scope of this chapter.
We also choose to omit a number of works, e.g.\ \citep{Longo_16,Zeng_18}, where ML is proposed as an application of an argumentative methodology only informally, without a concrete ML application. 
Further, approaches which consider tasks most often associated with ML methods but without some explicit use of ML, e.g.\ \citep{Amgoud_08,Potyka_22_ArgXAI}, or those which consider ML models as a possible source of knowledge, e.g.\ \citep{Hung_22}, are also omitted.
Similarly, we also choose to leave out methods which generate argument graphs from data that do not any include any explicit ML techniques, e.g.\ \citep{Hunter_20}. 
Orthogonally, we do not include works that target typical ML tasks and can be seen to have argumentative flavour but do not use argumentation in any way, e.g.\ \citep{Taniar_08}.
Finally, argumentative approaches which consider techniques often employed in ML, such as Markov decision processes (MDPs), e.g.\ \citep{Potyka_20}, or Bayesian networks (BNs), e.g.\ \citep{Bex_16}, but without explicitly using ML itself are typically outside the scope of the chapter.

\section{Literature Overview}
\label{sec:overview}

We start by overviewing research works that we deem to use argumentation for the benefit of ML (§\ref{subsec:arg}).
We then move on to overview works that we instead deem using ML mostly for the benefit of argumentation (§\ref{subsec:ml}).

\subsection{Argumentation for ML}
\label{subsec:clf}
\label{subsec:arg}

In this subsection, we consider the various ways in which argumentation has been used for supplementing ML models. 
Here, the general trend is for methods to target the improvement of the ML model's performance, explain its outputs or a combination of the two. 
We begin with those which concern improvement alone, before transitioning to those which concern explanation alone roughly according to the works' main objectives.
The works whose main objective is judged to be improvement in performance are discussed in §\ref{sssec:improving} and summarised in Table \ref{table:improving}, while those where the main objective is enhanced explainability are discussed in §\ref{sssection:explaining} and summarised in Table \ref{table:explaining}.\footnote{Note that in all tables we show only the references to the primary works describing an approach, with the others given in the discussion.}

\subsubsection{For Improving ML Models}
\label{sssec:improving}

\begin{table}[H]
\scriptsize
    \centering
    \begin{tabular}{lllllll} \hline
\!\!\!\!\textbf{Work} & 
\!\!\!\!\textbf{Type} & 
\textbf{Arg.} & 
\!\!\textbf{Learn.} &
\textbf{ML Model(s)} & 
\textbf{Task} & 
\!\!\!\!\textbf{Data} \\ \hline
\!\!\!\!\citep{Gomez_2004_JCST} & 
\!\!\!\!Seg & 
SA (DeLP) & 
\!\!U &
Fuzzy ART NN & 
Clust & 
\!\!\!\!Tab \\ 
\!\!\!\!\citep{Carstens_15} & 
\!\!\!\!Seg & 
QBAF & 
\!\!S &
NBC, SVM, RF & 
Class & 
\!\!\!\!Text \\ 
\!\!\!\!\citep{Ayoobi_22} & 
\!\!\!\!App & 
AF, QBAF & 
\!\!U, S &
Rule & 
Class & 
\!\!\!\!Tab \\ 
\!\!\!\!\citep{Thimm_17} & 
\!\!\!\!Seg & 
SA (DeLP) & 
\!\!U, S &
Rule & 
Class & 
\!\!\!\!Tab \\ 
\!\!\!\!\citep{Mozina_07} & 
\!\!\!\!Syn & 
Informal & 
\!\!S &
Rule & 
Class & 
\!\!\!\!Tab \\ 
\!\!\!\!\citep{Shao_20} & 
\!\!\!\!Syn & 
Informal & 
\!\!S &
NN & 
Class & 
\!\!\!\!Im\\ 
\!\!\!\!\multirow{2}{*}{\citep{Sendi_19}} & 
\!\!\!\!\multirow{2}{*}{Seg} & 
SA (RB, & 
\!\!\multirow{2}{*}{S} &
\multirow{2}{*}{NN, Rule, DT} & 
\multirow{2}{*}{Class} & 
\!\!\!\!\multirow{2}{*}{Tab}\\ 
& & Dialogues) & & & & \\

\!\!\!\!\citep{Prentzas_22} & 
\!\!\!\!Seg & 
SA (Gorgias) & 
\!\!S &
Rule, RF & 
Class & 
\!\!\!\!Tab \\ 
\!\!\!\!\citep{Riveret_20} & 
\!\!\!\!Syn & 
PA (Semi-AF) & 
\!\!S & 
RBM & 
Class & 
\!\!\!\!Tab \\ 
\!\!\!\!\citep{Gao_12} & 
\!\!\!\!Syn & 
VAF & 
\!\!R &
MDP & 
Policy & 
\!\!\!\!Sim \\ 
\!\!\!\!\citep{Riveret_19} & 
\!\!\!\!Syn & 
PA, SA & 
\!\!R &
MDP & 
Policy & 
\!\!\!\!Sim \\ 
\hline
    \end{tabular}
    \caption{A summary of the ways in which argumentation has been used for \emph{improving} ML models, as detailed in §\ref{sssec:improving}. We show: 
    the type of integration (\underline{seg}mented, \underline{syn}ergistic or \underline{app}roximate); 
    the kind of \underline{arg}umentation used; 
    the type of \underline{learn}ing employed (\underline{s}upervised, \underline{u}nsupervised or \underline{r}einforcement);
    the ML model(s) used, including 
    fuzzy adaptive resonance theory (ART) NNs, \underline{rule} learning (including rule induction and rule extraction) and restricted Boltzmann machines (RBMs); the task undertaken (\underline{clust}ering, \underline{class}ification or \underline{policy} learning) and the type of data used (\underline{tab}ular, 
    \underline{text}ual,
    \underline{im}age or, specifically in the case of RL, \underline{sim}ulation).}
    \label{tab:improving}
    \label{table:improving}
\end{table}

Several works propose either segmented, synergistic or approximate methods of integrating argumentation with ML models for the purpose of improving their performance. 
The general (but not exclusive) theme in these works is that of using rule learning to induce rule-based arguments and then to apply argumentative reasoning to 
improve the outputs of the ML models, 
typically using injected expert knowledge which is enabled by means of argumentation. 
We begin with some works which focus on the improvement of an ML model's performance exclusively, without considering explainability.

Some of the earliest works to use argumentation (as it is now commonly understood) for improving ML methods in classification tasks are those of \cite{Gomez_04, Gomez_2004_JCST}.
In particular, the authors propose the use of argumentation for \tags{tabular data clustering} using \tags{unsupervised learning}. 
In their \tags{segmented} approach, defeasible argumentation processes the outputs of an \tags{NN}. 
The use of argumentation is meant to improve on the 
then state-of-the-art of randomly choosing one output cluster when the input is pattern-matched to more than one cluster. 
In particular, data instances are first clustered using a \tags{fuzzy adaptive resonance theory NN}, and then \tags{DeLP} resolves inconsistent classification outcomes for instances assigned to multiple conflicting clusters. 
Specifically, the learnt clusters are represented in DeLP by means of facts and rules, and so are any preferences that would resolve situations where an instance is assigned to multiple clusters (for example, a cluster with newer instances is preferred over others, or the smallest/largest cluster is preferred). 
For a new data instance, DeLP-style argumentation finds the rules that are pattern-matched to apply and through dialectical analysis infers the preferred classification. 
The stated advantages of combining unsupervised clustering and preferences for argumentative conflict resolution are theoretical: the authors suggest their approach could be applied to resolve classification conflicts for ambiguous instances and to potentially explain classification decisions.

\cite{Carstens_15} define a \tags{segmented} approach for sentiment polarity \tags{classification} via \tags{supervised learning}. They leverage on the reasoning contained in \tags{QBAFs} with gradual semantics to make improvements in classifier accuracy in tasks with \tags{textual data}.
Here, the QBAF contains arguments representing the possible classes, as well as arguments comprising one or more premises and a conclusion, where premises are characteristics of sentences that indicate a certain polarity in the case of sentiment classification, and the conclusion is the polarity associated with these words. Base scores are obtained from the output of the classifier (e.g.\ confidence of the classification or the classification performance on the training corpus, depending on the classifier used) or from a given argument base. The dialectical strength of each class is computed using a quantitative semantics and the classification with maximal strength is assigned as the final classification for the testing instance.
The authors evaluate the approach on two corpora of tweets, the Sanders corpus and the STS corpus, and another corpus consisting of positive and negative sentences from movie reviews. 
They experiment with \tags{NBCs}, \tags{SVMs} and \tags{RFs}, showing improvements in accuracy in two computational linguistics tasks.

Going even further towards argumentation as an alternative to classical learning-based inference, 
\cite{Ayoobi_22} propose an \tags{approximate} method of pure argumentation-based learning for action selection in \tags{semi-supervised learning} scenarios.
The approach uses two \tags{argumentation frameworks}, one \tags{abstract} and one \tags{bipolar with weights} on relationships, to incrementally record -- as well as select the best -- actions given context described as sets of feature-value pairs. 
In short, arguments are of the precondition-postcondition form, obtained by \tags{rule learning} from
all the possible combinations of feature-value pairs of a data instance.
By default, the grounded extension of the abstract framework yields exactly one of the previously seen actions as the best in a given context and the supporting context conditions as feature-value pairs are recorded in a BAF. However, if the AF does not unambiguously resolve the conflicts among all the applicable actions, then an alternative action is extracted from the BAF depending on the cumulative strength of the already learnt support and attack conditions for various actions. 
The authors thus present a non-ML-based approach to \tags{classification} on \tags{tabular data}. 
Having in mind potential applicability to robotic scenarios, the authors compare this argumentation-based learning approach to some classic RL-type of approaches in multi-armed bandit problems as well as to incremental online learning approaches, and show superiority of argumentation-based learning with respect to both performance and speed of learning.
This work was extended in \citep{Ayoobiy_21}, which uses only BAFs with novel acceleration strategies thus addressing the complexity issues in \citep{Ayoobi_22} with better run-time, memory efficiency and learning accuracy.

The papers we overview next begin to consider how the incorporation of argumentation can bring benefits with regards to explainability, while still focusing on improving the ML model's performance as the main objective.

A short position paper \citep{Thimm_17} that rests on improving ML models by means of argumentation in the spirit of \cite{Gomez_04, Gomez_2004_JCST} summarises the use of \tags{SA} for improving either \tags{supervised} or \tags{unsupervised classification} of \tags{tabular data} with, crucially, the added benefit of explainability.
The authors propose a two-step, \tags{segmented} classification approach combining \tags{rule learning} and argumentative inference. 
First, rule learning algorithms extract from a given data set frequent patterns as lists of (typically) collectively conflicting rules. 
Then, such rules can be used to construct an SA framework from which conflict-free inferences can be drawn.
This method enables learnt and argumentatively supported classification of unseen data, purportedly while being explainable in the sense of showing why one class is preferred over others in terms of rule-based arguments. 
The authors show empirically how to avoid inconsistent classification via mined rules, making use of the \tags{DeLP} argumentative conflict resolution mechanism.

An early work that aims to improve \tags{classification} of \tags{tabular data} via \tags{supervised learning} using \tags{informal argumentation} is that of \cite{Mozina_07, Bratko_09}. The authors propose a \tags{synergistic} approach wherein experts provide arguments in the form of rules associating a class and feature-value pairs, in effect giving the machine explanations as feedback. The introduced approach first learns if-then rules from argumentative examples by extending an existing method for \tags{rule learning}, the CN2 rule induction algorithm. Two types of arguments are incorporated: "C because Reasons" and "C despite Reasons", where the first type provides reasons (i.e.\ combinations of features) for why a certain training instance is classified as is, whereas the second type highlights combinations of features that do not play a role in the classification of a training instance. 
Importantly, incorporating arguments in this way allows experts to provide information which constrains the ML model's training, facilitating interactive ML between users and models, as described in a later work \citep{Mozina_18}.
The authors evaluate the approach empirically with three datasets from the UCI machine learning repository, and exemplify the approach in the legal and medical domains.

Another example of using \tags{informal argumentation} in a \tags{synergistic} way for improving \tags{classification} via \tags{supervised learning}, but with \tags{image data}, is \citep{Shao_20}. 
The authors suggest that a probabilistic neuro-symbolic image classifier can be argued with by a human user about the correctness of classification.
In contrast to previous works discussed in this subsection, they use influence functions to expose saliency maps as explanations alongside the instances classified by an \textbf{NN} (an 8-layer multi-layer perceptron (MLP)). 
The user can then argue with those outputs by putting forward (automatically generated) counterexample images to correct classifications or directly regularising the gradients to penalise incorrect explanations.
The authors show experimentally on three image datasets that their approach can improve classifier accuracy on both new examples and previously misclassified counterexamples.

Using argumentation more formally in a \tags{segmented} way, \cite{Sendi_19} create argumentative classifiers from rules extracted from \tags{NNs} for the purposes of injection of domain knowledge and more explainable \tags{classification} of \tags{tabular data} via \tags{supervised learning}. 
Specifically, an NN model trained on tabular data (using various UCI datasets) is approximated into a more interpretable model in two ways: either as a \tags{DT} (using TREPAN library by \citeauthor{DBLP:conf/nips/CravenS95}~(\citeyear{DBLP:conf/nips/CravenS95})) or by eclectic \tags{rule extraction}. 
The rules (which amount to paths in a DT) constitute \tags{premise-conclusion type of arguments}, where premises amount to feature-value pairs and the conclusion to a class label together with the classifier's confidence score. 
These rules are said to essentially function as explanations for the classifier's outputs.
Additionally, experts' arguments are also modelled in the premises-conclusion form. 
This allows for a segmented approach of an approximated NN classifier and expert knowledge into an argumentative classifier. 
The authors suggest how multiple argumentative classifiers can engage in multi-agent argumentation by means of a \tags{dialogue protocol}, whereby conflicts among arguments with different conclusions are resolved by prioritising first expert knowledge arguments and then the ones with higher confidence scores. 
They provide a case study of multi-class medical treatment recommendation task in a realistic virtual population of 40,000 individuals (datapoints) described using numerous features, and further experiment with 11 UCI binary classification datasets comparing their method to ensemble learning and rule extraction methods (such as boosted trees and ensemble NNs).  
Experimental results indicate that such multi-agent argumentative classifiers (albeit without expert agents) outperform ensemble methods based on either the original or baseline NN models in terms of classification accuracy. 
The authors also posit that their approach enhances explainability by exposing the outputs of NNs with argumentative resolution.

\cite{Prentzas_22} define a \textbf{segmented} approach to argumentation-based \tags{classification} of \tags{tabular data} via \tags{supervised learning}, with an application in the field of cancer prognosis. 
The authors utilise \textbf{SA}, in the form of the \tags{Gorgias argumentation framework}, where the basic arguments consist of premises and a conclusion, and other arguments express preferences over the basic arguments.
After undertaking some pre-processing in the form of statistical analyses of features' significance, the authors prescribe some form of \tags{rule extraction} method to create arguments, using \tags{RF} as an example. 
Next, in an iterative and manual process, tailored defeat arguments for rules are defined, before the dilemmas caused by argument conflicts are reduced, by adding arguments which adjust the preferences.
The authors then deploy the approach in cancer prognosis prediction, demonstrating that it exhibits reasonable accuracy in an empirical analysis before showing the natural explainability afforded by its argumentative nature. 
This explainability is explored further, with varying styles of explanation, via a web platform in the social media, legal and medical domains in \citep{DBLP:conf/comma/SpanoudakisKK22}.

\cite{Riveret_20} propose a neuro-symbolic method to \tags{supervised learning} that \textbf{synergistically} combines graphical representation, probabilistic learning and argumentative reasoning. 
Their aim is to address the challenges of learning from examples the probabilistic statuses and dependencies among arguments/statements in PA, 
and that of constraining and explaining probabilistic inferences of NNs for \tags{classification}. 
To this end, they integrate \textbf{RBMs} and \textbf{probabilistic semi-abstract argumentation} in the following way. 
First, they represent each datapoint in a set of \tags{tabular data} as an argument graph with nodes being feature-value pairs, along with the class label, and edges encoding prior knowledge in the form of logical constraints about the dataset, so that the constraints are captured by attack and support relationships and an argument labelling includes classification of the datapoint. 
Second, prior knowledge constraints as argument labellings are integrated within RBMs, so that the latter are used to learn and respect the probabilistic dependencies amongst argument labels. 
Finally, formulating the learning task for binary classification over datapoints as argument labellings yields neuro-symbolic argumentation machines (NSAMs) -- a probabilistic learning and reasoning method as an alternative to the standard RBMs as well as other ML methods. 
In this way, NSAMs can learn argument graphs from data and use argumentation to help with both classification of aberrant datapoints in noisy settings and overfitting. 
The authors compare NSAMs with classic ML approaches such as NNs, logistic regression (LR) and DTs, among others. 
They contrast in particular with the standard RBMs that only learn from data with those that use both prior knowledge constraints and learn from data, namely NSAMs.
An experiment on a small (2400 examples) tabular dataset augmented with noise and for which expert knowledge is available shows that NSAMs outperforms other methods. 
As an added benefit, the authors claim that NSAMs provide explanations of individual classifications in terms of maximally consistent argument graph labellings that pertain to the pertinent feature-value pairs and prior knowledge constraints.

The works we have overviewed here focus on improving and explaining ML-based classification by means of typically a form of SA. 
The next family of works that we discuss focuses on RL instead. Concretely, the 
works aim at improving RL-based agent policies in Multi-Agent Systems (MASs).

The use of argumentation to improve RL-based agents and their interactions generally follows a synergistic approach of introducing argumentative reasoning about the utilities of actions recommended by an RL agent's policy. 
For example, \cite{Gao_12} develop a \textbf{synergistic} method of argumentation-based 
\tags{RL}, where domain expert knowledge is injected into a (semi-)\tags{MDP}-based algorithm to use as a reward shaping technique by means of \tags{VAFs}.
The authors specifically show how an on-policy learning algorithm SARSA can be modified to use an argumentation-based look-ahead reward shaping function using \tags{simulations} in the application of RoboCup Soccer \citep{Robocup}. 
In particular, domain-knowledge (about soccer) is used to define VAFs, where values roughly represent different tactics given by domain experts, so that numerical rewards are assigned to an RL agent's actions based on the arguments found in the preferred extensions of the framework. 
Experimental results show that the performance of thusly modified RL algorithm can significantly improve \tags{policies} relative to those of the original. 
\cite{Gao_14} extend this work to the multi-agent RL setting.

Further work by some of the same authors in \citep{Riveret_19} combines \tags{RL} with PA instead, again in a \tags{synergistic} manner. 
Argument values in \tags{probabilistic SA frameworks} are quantified as utilities and learnt using RL. Roughly, \tags{MDPs} are captured as argumentation-based agents using \tags{SA} frameworks with \tags{temporal modality operators} and a \tags{probability distribution} over argument labellings. Reinforcement is  modelled as probabilistic update of argument labels in time. One of the main benefits of this approach is that agents can have attitudes (expressed as MDP actions) and thereby deliberate about \tags{policies} using the underlying logic and argument labellings. The authors mention the advantage of the natural agency afforded by argumentation in the form of an ability to explain and forecast agents' actions,
and illustrate their points with experiments in an environment simulating a simple hand-crafted MDP.

While the above works purport to show improvements of 
ML models by means of argumentation in specific scenarios, the benefits typically rely on the availability of user/expert interaction, e.g.\ by means of adding knowledge in terms of rules or preferences. 
It is also not clear whether or not the state-of-the-art ML has already moved beyond the problems addressed in these works.

Though these discussed works generally aim at improving ML models' performance with argumentation, most of them also suggest that argumentation can also help to explain the inferences. 
In the next section we discuss a large body of works in which the primary goal is that of using argumentation for explaining ML models.

\subsubsection{For Explaining ML Models}
\label{sssection:explaining}

\begin{table}[H]
\scriptsize
    \centering
    \begin{tabular}{lllllll} \hline
\!\!\!\!\textbf{Work} & 
\!\!\!\!\textbf{Type} & 
\textbf{Arg.} & 
\!\!\textbf{Learn.} &
\!\!\!\!\textbf{ML Model(s)} & 
\!\!\!\!\textbf{Task} & 
\!\!\!\!\textbf{Data} \\ \hline
\!\!\!\!\multirow{3}{*}{\citep{Cocarascu_20}} & 
\!\!\!\!\multirow{3}{*}{Seg/Syn} & 
\multirow{3}{*}{AF} & 
\!\!\multirow{3}{*}{U, S} &
\!\!\!\!\multirow{3}{*}{AE, RF} & 
\!\!\!\!\multirow{3}{*}{Class} & 
\!\!\!\!Tab, 
\\ 
& & & & & & \!\!\!\!Text,  \\
& & & & & & \!\!\!\!Im \\
\!\!\!\!\multirow{2}{*}{\citep{Prakken_22}} & 
\!\!\!\!\multirow{2}{*}{Seg/Syn} & 
\multirow{2}{*}{AF} & 
\!\!\multirow{2}{*}{S} &
\!\!\!\!DT, SVM, & 
\!\!\!\!\multirow{2}{*}{Class} & 
\!\!\!\!\multirow{2}{*}{Tab}
\\ 
& & & & \!\!\!\!NBC, LR & & \\
\!\!\!\!\multirow{2}{*}{\citep{Lente_22}} & 
\!\!\!\!\multirow{2}{*}{Syn/App} & 
\multirow{2}{*}{SA} & 
\!\!\multirow{2}{*}{S} & 
\!\!\!\!LR, SVM, & 
\!\!\!\!\multirow{2}{*}{Class} & 
\!\!\!\!\multirow{2}{*}{Tab} \\ 
& & & & \!\!\!\!RF, NN, RE & & \\
\!\!\!\!\citep{Bistarelli_22} & 
\!\!\!\!App & 
BAF & 
\!\!S & 
\!\!\!\!BB & 
\!\!\!\!Class & 
\!\!\!\!Tab \\ 
\!\!\!\!\citep{Vilone_22} & 
\!\!\!\!Seg/App & 
SA (RB) & 
\!\!S &
\!\!\!\!NN & 
\!\!\!\!Class & 
\!\!\!\!Tab \\ 
\!\!\!\!\citep{Kazhdan_20} & 
\!\!\!\!App & 
VAF & 
\!\!R & 
\!\!\!\!TE & 
\!\!\!\!Policy & 
\!\!\!\!Sim \\ 
\!\!\!\!\citep{Otero_23} & 
\!\!\!\!App & 
VAF & 
\!\!R &
\!\!\!\!MDP, AD & 
\!\!\!\!Policy  & 
\!\!\!\!Sim \\ 
\!\!\!\!\citep{Lertvittayakumjorn_23} & 
\!\!\!\!Syn & 
QBAF & 
\!\!S & 
\!\!\!\!LR & 
\!\!\!\!Class &
\!\!\!\!Text \\
\!\!\!\!\multirow{2}{*}{\citep{Timmer_17}} & 
\!\!\!\!\multirow{2}{*}{App} & 
SA & 
\!\!\multirow{2}{*}{S} &
\!\!\!\!\multirow{2}{*}{BN} & 
\!\!\!\!\multirow{2}{*}{Class} & 
\!\!\!\!\multirow{2}{*}{Tab} \\ 
& & (ASPIC$^+$) & & & & \\
\!\!\!\!\citep{Dejl_21} & 
\!\!\!\!App & 
GAF & 
\!\!S &
\!\!\!\!NN & 
\!\!\!\!Class & 
\!\!\!\!Text \\ 
\!\!\!\!\citep{Sukpanichnant_21} & 
\!\!\!\!App & 
QBAF & 
\!\!S &
\!\!\!\!NN & 
\!\!\!\!Class & 
\!\!\!\!Im \\ 
\!\!\!\!\citep{Mollas_22} & 
\!\!\!\!Seg/App & 
SA &
\!\!S &
\!\!\!\!BB & 
\!\!\!\!Class & 
\!\!\!\!Tab \\ 
\!\!\!\!\citep{Amgoud_23} & 
\!\!\!\!Seg & 
AF & 
\!\!S &
\!\!\!\!BB & 
\!\!\!\!Class & 
\!\!\!\!Tab \\ 
\!\!\!\!\citep{Potyka_23} & 
\!\!\!\!App & 
BAF & 
\!\!S &
\!\!\!\!RF & 
\!\!\!\!Class & 
\!\!\!\!Tab 
\\
\hline
    \end{tabular}
    \caption{A summary of the ways in which argumentation has been used for \emph{explaining} ML models, as detailed in §\ref{sssection:explaining}. We show: 
    the type (\underline{seg}mented, \underline{syn}ergistic or \underline{app}roximate);
    the kind of \underline{arg}umentation used; 
    the type of \underline{learn}ing employed (\underline{s}upervised, \underline{u}nsupervised or \underline{r}einforcement);
    the ML model(s) used, including 
    autoencoders (AEs), rule extraction (RE), 
    trajectory extraction (TE), autoregressive decoding (AD) and black-boxes (BBs); the task undertaken (\underline{class}ification or \underline{policy} learning) and the type of data used (\underline{tab}ular, 
    \underline{text}ual,
    \underline{im}age or \underline{sim}ulation).}
    \label{tab:explaining}
    \label{table:explaining}
\end{table}

We now consider approaches which incorporate argumentation to enhance explainability as the main objective, firstly considering approaches which still target improvement in the ML model's performance. One point of note here is the increase in the number of methods which are approximate, in line with the well known trend of explanations being simplifications of the original models \citep{Rudin_19}, such that they are cognitively manageable for humans.

\cite{Cocarascu_20} take a \textbf{segmented}/\textbf{synergistic} approach to develop a model for \tags{classification} of \tags{tabular}, annotated \tags{image} and \tags{textual data} by leveraging ML's capability to compactly represent data and the argumentative reasoning to make and explain decisions. The work integrates simple, classic \tags{supervised} and \tags{unsupervised learning} techniques for data representation with \textbf{AFs} for relating data points and reasoning about their labels. 
In detail, an \textbf{autoencoder} (a type of NN used to learn efficient encodings of unlabelled data) is used to reduce the dimensionality of tabular (Mushroom dataset with 126 one-hot encoded features) and annotated image data (CelebFaces Attributes and Objects with Attributes datasets with, respectively, 40 binary attributes, and 64 attribute labels divided into 20 classes) so that data instances could be more compactly represented as arguments in an AF. 
In the case of textual data, \tags{RFs} are trained on one-hot-encoded vocabularies of (possibly semantically-clustered) word lemmas extracted from product reviews with positive and negative sentiments, using two movie review datasets. The outputs of RFs serve as weights on words so that arguments would represent datapoints as sentiment-labeled sets of weighted words.  
The AFs are inspired by case-based reasoning (CBR), whereby an informativeness relation (over subsets of (weighted) features) together with datapoint label divergence are used to  define attacks among arguments/datapoints. Further, a default label and argument are designated, and the reasoning amounts to deciding whether the default argument is defended by the grounded extension of the AF augmented with an unlabeled datapoint/argument. 
The determination of belonging to the grounded extension amounts to  classification of the unlabeled datapoint. 
Finally, the reasoning about the acceptability of the default argument according to the grounded semantics is presented as an argumentation debate for the purpose of explaining the classification of a given datapoint in terms of its relations to other labeled datapoints. 
In this way, the overall framework first makes use of ML for data representation and then replaces a standard ML way to classification with that of argumentative inference (and explanation) over compactly represented data.

\cite{Prakken_22} take the CBR-inspired argumentation for explainable \textbf{classification}, in the spirit of \citep{Cocarascu_20}, further. 
The authors consider an arbitrary \textbf{supervised} classifier trained on \textbf{tabular data} and build an \textbf{AF}-based model that justifies the outcomes of the classifier in terms of feature-value pairs and/or relevant datapoints. 
The \textbf{segmented/synergistic} approach assumes access to data features and the training set. The method incorporates rationale from legal reasoning: basically, data features (known as factors) and datapoints (known as cases) tend to favour one classification outcome or another. 
The authors employ a dialogical exchange of arguments (in terms of factors and cases) for and against a particular classification as an explanation mechanism. 
They experiment with classic ML models, including \tags{DTs}, \tags{SVMs}, \tags{NBCs} and \tags{LR}, 
on three basic tabular datasets (Mushroom, Churn, Admission). 
The results show, among other things, that the dialogical argument exchange can indeed explain (justify) most of the ML model's correctly classified datapoints.

\cite{Lente_22} introduce argumentative explanations, which are said to be modelled on those used by humans in everyday life, for any \tags{supervised} black-box \tags{classification} model operating on \textbf{tabular data}. The method is model-agnostic because it deals with inputs and outputs of the classifier, learning \tags{SA frameworks} in a \textbf{synergistic} manner via a \textbf{rule learning} algorithm such that they include features as premises, classes as conclusions and a quantitative strength value of the argument. 
Then, attacks between arguments are drawn when an argument rebuts another, and the arguments are then evaluated using the grounded semantics. 
This argumentation framework can then be used for the classification task itself, thus \tags{approximating} the ML model, but also for explaining it, where explanations are selected as subgraphs of the argument graph, such that the arguments are either pro and con the classification. 
In the experiments, the authors demonstrate how the explanations perform with respect to five different evaluation metrics, notably fidelity, i.e.\ how well the explanations approximate the prediction of the ML model, and accuracy, i.e.\ how well the argumentation framework predicts with unseen data. They experiment with four different types of ML model, namely \tags{LR}, \tags{SVMs}, \tags{RFs} and \tags{NNs}, with datasets from the UCI ML Repository.

Another method which introduces a technique for \tags{approximating} \tags{black-box} ML models for \tags{supervised classification} tasks with argumentation frameworks is that of \cite{Bistarelli_22}. 
In an extended abstract, the authors propose to approximate ML-based classifiers operating on \tags{tabular data} using \tags{BAFs} with arguments as feature-value pairs in favour or against a particular class, whence sub-trees with arguments accepted or rejected under semi-stable semantics serve as model- and data-agnostic explanations of the classification. 
In particular, the authors first cluster the dataset based on ranges of the values of variables, before they compute a correlation matrix to approximate relations the features. 
These relations are initially symmetrical but the authors use conditional probabilities to remove some edges, while ensuring that the graph remains connected, and determine which arguments attack and support one another.
The authors then show how, given a dataset, the semi-stable semantics for BAFs can be used to generate an explanation for given classifications, approximating the classifier via the dataset.

\cite{Vilone_22} also define a model-agnostic method for \tags{approximating} ML models undertaking \tags{supervised classification} on \tags{tabular data} in a \textbf{segmented} manner, this time deploying \textbf{rule-based SA} frameworks with the ranking-based categoriser semantics. This is achieved by defining arguments as IF-THEN rules, where the premises and conclusion of an argument correspond to the rule's antecedents and conclusion. 
The premises consist of restrictions of the features' values, while the conclusions are classes predicted by the classifiers, thus resulting in global explanations.
Attacks between arguments are then extracted based on rebutting and undercutting, which are weighted and pruned based on the coverage in the dataset of the corresponding rules. 
Finally, the arguments are evaluated by the aforementioned semantics, with the acceptable arguments serving as explanations for the classifier. 
These explanations are  evaluated using feed-forward \tags{NNs} trained on 5 datasets from Kaggle 
and the UCI ML Repository, where they are shown to be simpler and more comprehensible than DTs, but lack their faithfulness, correctness and robustness.

As in §\ref{sssec:improving} where we considered several works that focus on argumentation improving RL, we next discuss a couple of works which focus on explaining as well as improving RL.

\cite{Kazhdan_20} extract models from multi-agent \tags{RL} systems and augments them with expert knowledge-based arguments for explanatory purposes. Specifically, \tags{model extraction} (in some form) from RL agents' \tags{trajectories} (i.e.\ sequences of state-action pairs) as recorded in data logs is assumed. Then arguments corresponding to recommendations of actions to perform in a given state (for any one agent) are assumed to be given by an expert user, and numerical values are assigned to those arguments based on the frequency of the actions taken in the corresponding states as observed in the trajectories. 
A \tags{VAF} is used to represent the user-provided arguments together with values obtained from data, and grounded semantics is used to define \tags{policies} selecting the best actions. An argumentation-based model of MARL agents is thus constructed in a way that attempts to both \tags{approximate} the agents by expert heuristics, and enables expert knowledge to be incorporated. 
The effectiveness and interpretability of thusly extracted models is illustrated with \tags{simulation data} in a RoboCup Soccer \citep{Robocup} scenario.

\cite{Otero_23} note that in the setting of multi-agent \tags{RL}, non-symbolic models aided by argumentation can learn high-performing \tags{policies}, as in \citep{Gao_13,Gao_14}, but are hard to explain due to their non-symbolic nature. 
They likewise note that while model-extraction methods using argumentation provide explainability to non-symbolic models by producing symbolic counterparts, as in \citep{Kazhdan_20}, they may incur performance loss and are often approximate and thus not completely faithful to the underlying non-symbolic model. The authors thus aim to directly learn a symbolic argumentation-based multi-agent RL model that, while still \tags{approximate}, is both performant and explainable. 
The authors use RL to preference-rank arguments in an expert-constructed \tags{VAF} in a way that maximises the performance score of the VAF, i.e.\ how good the VAF is a reasoning engine for choosing actions in some task. 
More concretely, they first formulate a combinatorial optimisation problem as an \tags{MDP} with the state space as the set of all the possible partial rankings over arguments and the action space as arguments themselves. 
Where a state is a total ranking, it is assigned a score (i.e.\ reward) by executing the reasoning in a VAF with the arguments ranked in preference accordingly to the state. 
Otherwise, states representing partial argument orderings are completed by an RL algorithm step by step: it learns to add arguments until a total ordering is reached by maximising the reward given for a total argument ordering and updating the policy-gradient weights accordingly (this is also know as \textbf{autoregressive decoding} in combinatorial optimisation via RL).
The learnt VAF can be used as an agent's policy recommending actions promoted by the accepted arguments. 
The authors implement and compare their approach with some symbolic model-based agent architectures, including those from \citep{Kazhdan_20}, with \tags{simulation data} in the multi-agent RL environment RoboCup Keepaway \citep{Robocup}. They establish their approach to be superior both performance- and explainability-wise, where in the latter case, the best policy VAF learnt with the help of RL is completely faithful to the trajectories of execution, in contrast to only approximate fidelity achieved in \citep{Kazhdan_20}. 
Overall, the approach produces performant argumentation-based agent models that can be verified and yield faithful explanations of the agent's bahaviour in a multi-agent RL setting.

We now consider the approaches where explainability of the ML model is the sole focus of the works, rather than improving its performance.

In \citep{Lertvittayakumjorn_23}, the authors target argumentative explanations of pattern-based \tags{LR} (PLR) for binary \tags{text} \textbf{classification} via \tags{supervised learning}. They show that the interpretability of PLR does not guarantee that explanations are amenable to humans, demonstrating that standard explanation methods may be seen as being implausible. To rectify this, the authors introduce argumentative explanations which are able to provide additional layers of explainability, clearly demonstrating relationships between the patterns, such as agreement or disagreement. 
The \tags{synergistic} method uses \tags{QBAFs} as a means to represent the patterns observed by the classifier in an approximate manner, with the gradual semantics over the QBAF modelling the LR. They show that, given an input, the classifier and its QBAF representation always give the same prediction, and prove intuitive argumentative properties of the QBAF.
The method is then evaluated empirically with three datasets,  
proving its advantages wrt sufficiency, and experimentally in two user studies with humans, showing improved plausibility and helpfulness over other explanation methods.

\cite{Timmer_17} propose \tags{SA} to \tags{approximate} and explain reasoning in \tags{BNs} performing \tags{supervised classification} with \tags{tabular data}. First, for a variable of interest in a BN, a support graph is extracted, consisting of chains of variables whose observations will propagate in the BN and influence the variable of interest. Following the relations in a support graph (which is independent of the observed evidence),  \tags{ASPIC$^+$}-like arguments with premises and conclusions as variable-outcome pairs are constructed, and their strengths as either likelihood ratio or posterior odds are calculated from the premises given observed evidence.  
Conflicts among arguments are resolved using their strengths as priorities and the grounded semantics captures arguments with highest probabilities. 
This argumentative approximation helps explain the conclusions drawn from a BN, as illustrated in a legal reasoning scenario.

\cite{Dejl_21} implement a method for generating argumentation frameworks which \tags{approximately} represent various forms of \tags{NNs}, trained in a \tags{supervised} manner for \tags{classification} with \tags{textual data}, and provide customisable explanations for their outputs. 
This method represents neurons and connections as arguments and relations, respectively, in a \textbf{generalised argumentation framework}.
These relations are extracted when the connections between two neurons satisfy certain conditions, and so the argumentation framework highlights parts of the network which may be interesting to a user as part of an explanation. 
The explanations themselves are provided in a variety of different forms, e.g.\ visual and conversational, which renders them suitable for a number of purposes.
In the paper, the method is exemplified in the context of text classification, but other tasks such as image classification are supported, e.g.\ see \citep{Albini_20X}.

A similar \tags{approximate} method is proposed by \citep{Sukpanichnant_21}, where we again have argumentation frameworks representing \tags{NNs} trained via \tags{supervised learning}, but this time using \textbf{QBAFs} specifically and targeting \tags{image classification} tasks. 
Arguments may represent sets of neurons in a layer, alleviating the density of the argumentation framework and thus extracting information which is more cognitively manageable to users. 
The authors show that existing methods for explanation in NNs may be represented by the QBAF, and then use it to provide visual explanations in the form of~\citep{Dejl_21}.

\cite{Mollas_22} \tags{approximate} \tags{black-box} ML models for \tags{classification} which operate on \tags{tabular data} in a \tags{segmented} manner with \tags{SA}, producing explanations which are claimed to be more truthful than standard feature importance explanations. The method works by checking, for each feature, whether feature importance measures are in line with expectations when perturbations occur in their values. 
Depending on these checks, trees  comprising structured arguments are constructed from fixed templates, which use both rebutting and undercutting attacks to determine whether the explanation is truthful, i.e.\ aligns with the expectations for all features.  
The authors then perform experiments with standard datasets 
from the literature. Here, they assess the method both qualitatively, showing intuitive examples of the resulting explanations, and quantitatively, showing improvements over standard feature attribution explanations in truthfulness.

In \citep{Amgoud_23}, the authors first show that abductive explanations, i.e.\ those consisting of features and assigned values considered to be sufficient for a given class, can be guaranteed to satisfy either, but not both, existence or correctness, properties defined in previous work \citep{Amgoud_22}. 
They then introduce a parameterised family of argumentation-based explanation functions, which explain \tags{black-box} models for \tags{supervised classification} of \tags{tabular data} in a \tags{segmented} manner. Here, \tags{AFs} comprise arguments, consisting of features and assigned values which support classes, and attacks, which are identified between the arguments. 
The functions then use the stable semantics to select sets of arguments which are jointly considered to be acceptable.
These explanations are shown, using theoretical and empirical analyses, to not only guarantee correctness but explain a reasonable proportion of instances, i.e.\ performing well wrt existence.

\cite{Potyka_23} \tags{approximate} \tags{RFs} deployed as \tags{supervised classification} models of \tags{tabular data} as argumentation frameworks, specifically \tags{BAFs}, in order to provide global explanations for their reasoning.
The BAFs themselves consist of arguments representing the classes, the rules and subsets of the features' values. 
Attacks (occurring between features, from features to rules and from rules to classes) and supports (occurring from rules to classes only) are then drawn between these arguments.
It is then shown that extensions of the bi-stable semantics correspond to possible classification decisions.
This means that finding sufficient and necessary explanations for the classifications can be reduced to finding those in the BAFs.
The authors use Markov network encodings of the BAFs to solve the resulting combinatorial reasoning problems, and introduce an efficient probabilistic algorithm for approximating their solutions, given their high computational complexity.
Finally, the approach's ability for finding sufficient and necessary reasons is examined empirically using three datasets (Iris, PIMA and Mushroom) 
from the literature.

\subsubsection{Discussion}

Our review so far of the works where argumentation has been used for supplementing ML models allows us to make the first following conclusions.

\textit{Argumentative methods for improving ML-based inference show promise for small models and datasets, but it remains to be seen whether this can be replicated at scale.}
Argumentation has been proposed to improve ML-based inference, under the assumption that argumentation-based inference is better-suited to reasoning with rules of thumb and exceptions than ML-based pattern matching is, or that it can readily incorporate expert knowledge.
This is perhaps borne out in settings of small models and datasets, where patterns or rules discovered by ML can be augmented with, say, rule-based, argumentative reasoning.
Whether argumentation can be similarly helpful with modern large scale models and datasets remains to be seen, especially because modern ML inference resolves conflicts probabilistically (the more likely patterns yield preferred inference) and large scale datasets tend to allow for capturing exceptions and particular contexts using larger models. 
We hope though that argumentation (or similar symbolic reasoning methods, for that matter) could be useful in specific applications, especially those requiring injection of expert knowledge, but one would still need to show that such methods would be better than the current ML methods such as Federated Mixture of Experts (ensembles of specialized models) or RLHF (RL with Human Feedback).
However, it is not beyond the scope of the imagination that argumentation could be used to supplement these processes themselves, e.g.\ to resolve the conflicts between experts in the former case, as in \citep{Abchiche-Mimouni_23} for conflicting models in an ensemble, or to structure human feedback in RLHF, as in \citep{Rago_21} for recommender systems.
Thus, argumentation's usefulness in improving ML inference at scale remains to be seen, though there has been clear progress which has highlighted potential opportunities.

\textit{There is arguably more evidence for argumentation's suitability for supporting the explainability of ML models, a goal which does not seem to conflict with that of model improvement, though challenges remain.}
There are many diverse works which construct argument graphs highlighting features, rules, data clusters, or model components that provide reasons for particular predictions. 
These approaches are said to provide benefits with regards to improving model inference at varying degrees, with a full spectrum of improvement to explanation on display.
The intuitive relationship-based structure of argumentation explanations is seen as being more amenable to human understanding compared to typical attributive explanations (such as feature importance measures) used in ML. 
This seems promising, but widely-employed and convincingly useful systems for ML model explainability via argumentation are still somewhat lacking. 
Similarly, explainability of RL-based agents using argumentation seems promising, but much more work is needed to show its benefits in realistic systems that operate in high complexity and extensive simulation scenarios. 
Overall, engineering useful argumentation frameworks for explainability poses knowledge acquisition and scale bottlenecks. Open challenges also remain regarding explanatory power -- whether argumentation actually improves user trust or model transparency over state-of-the-art approaches. More studies using objective metrics and subjective experimental evaluation with users would be welcome, and, without this, 
the extent of argumentation's explanatory benefits may continue 
to be seen as somewhat speculative.

While we discussed the two broad categories of the use of argumentation in tandem with ML for the sake of structured prediction and multi-agent interaction in the previous sections, 
in the next section we overview how ML has been considered to support argumentation.

\subsection{ML for Argumentation}
\label{subsec:ml}

In this subsection, we first discuss some proposals for using argumentation frameworks for representing patterns learnt from data and making inferences thereon, essentially as a paradigm of argumentation-based learning to replace standard methods of supervised learning. 
We then overview several works that instead use ML to improve argumentation,  by either supporting learning of argumentation frameworks from data or suggesting to learn and predict argument acceptance statuses in argumentation frameworks. 
Next, we cover works which combine ML and argumentation for analysis of the two formalisms, with some overlap with the sections covered in §\ref{subsec:arg}.
We lastly cover a couple of works that focus on human dialogues in the real world and suggest that ML is perhaps more useful there than argumentation, implying that ML could be used in place of argumentation for some specific tasks.

\subsubsection{For Supporting Argumentation}
\label{sssec:mlsuparg}

\begin{table}[H]
\scriptsize
    \centering
    \begin{tabular}{lllllll} \hline
\!\!\!\!\textbf{Work} & 
\!\!\!\!\textbf{Type} & 
\textbf{Arg.} & 
\!\!\!\!\textbf{Learn.}\!\!\!\! &
\textbf{ML Model(s)} & 
\textbf{Task} & 
\!\!\!\!\textbf{Data} \\ \hline
\!\!\!\!\citep{Ontanon_12} & 
\!\!\!\!Seg & 
SA (RB) & 
\!\!\!\!S &
Inductive ML, Rule & 
Gen & 
\!\!\!\!Tab \\
\!\!\!\!\citep{DeAngelis_23} & 
\!\!\!\!Seg & 
SA (ABA) & 
\!\!\!\!S & 
Inductive ML, Rule & 
Gen & 
\!\!\!\!Tab \\ 
\!\!\!\!\citep{Cocarascu_19} & 
\!\!\!\!Seg & 
QBAF & 
\!\!\!\!S & 
LSTM, TM & 
Gen & 
\!\!\!\!Text \\ 
\!\!\!\!\citep{Craandijk_22} & 
\!\!\!\!App & 
AF & 
\!\!\!\!R & 
Q-learning with GNN & 
Enf & 
\!\!\!\!AF \\ 
\!\!\!\!\citep{Garcez_05} & 
\!\!\!\!App & 
VAF & 
\!\!\!\!S & 
NN (C-ILP) 
& 
Comp & 
\!\!\!\!Tab \\ 
\!\!\!\!\citep{Kuhlmann_19} & 
\!\!\!\!App & 
AF & 
\!\!\!\!S & 
GNN & 
Comp & 
\!\!\!\!AF \\
\!\!\!\!\citep{Malmqvist_20} & 
\!\!\!\!App & 
AF & 
\!\!\!\!S & 
GNN & 
Comp & 
\!\!\!\!AF \\ 
\!\!\!\!\citep{Craandijk_20} & 
\!\!\!\!App & 
AF & 
\!\!\!\!S &
GNN & 
Comp & 
\!\!\!\!AF  \\ 
\!\!\!\!\multirow{2}{*}{\citep{Klein_22}} & 
\!\!\!\!\multirow{2}{*}{App} & 
\multirow{2}{*}{AF} & 
\!\!\!\!\multirow{2}{*}{S} & 
kNN, NBC, RF & 
\multirow{2}{*}{Comp} & 
\!\!\!\!\multirow{2}{*}{AF} \\ 
& & & & SVM, GNN & & \\
\hline
    \end{tabular}
    \caption{A summary of the ways in which ML has been used for \emph{supporting} argumentation, as detailed in §\ref{sssec:mlsuparg}. We show: 
    the type (\underline{seg}mented or \underline{app}roximate); 
    the kind of \underline{arg}umentation used; 
    the type of \underline{learn}ing employed (\underline{s}upervised or \underline{r}einforcement);
    the ML model(s) used, including 
    \underline{rule} learning, 
    long-short term memory networks (LSTMs), 
    topic modelling (TM), k-nearest neighbours (kNN); the task undertaken, i.e.\ the purpose of using ML (argument graph \underline{gen}eration, \underline{enf}orcement or \underline{comp}utation of semantics and/or argument acceptance) and the type of data used (\underline{tab}ular, 
    \underline{text}ual or AFs).
    }
    \label{tab:supporting}
    \label{table:supporting}
\end{table}

One trend of using ML for supporting argumentation is that of using supervised learning to generate argumentation frameworks. 
We first discuss several works that consider inductive ML (also known as concept learning, which essentially is learning from positive and negative examples to generate IF-THEN rules for classification) as a method for inducing rules and SA frameworks thereof. 
We suggest the reader to follow up with the references cited in the discussed works to explore the potential connections of argumentation and inductive ML.

\cite{Ontanon_12} argue that inductive generalisation in ML is a form of defeasible reasoning, and define a logical model to characterise and \tags{generate} hypotheses from sets of examples from \tags{tabular data} in a \tags{supervised} manner. 
The authors then show how their logic-based learning model can be integrated with argumentation in a \tags{segmented} manner to allow for argumentative interaction between agents that have inductively learnt different theories.
The agents' theories are built from the rules learnt in the standard \tags{inductive ML} setup, and \tags{rule-based} argumentation (under, basically, grounded semantics) is used to reconcile the theories that different agents hold to be mutually consistent.

More recently, \cite{Proietti_23X,DeAngelis_23} have proposed logic-based learning of \textbf{SA} frameworks from positive and negative examples in \tags{tabular data}. The authors consider settings where examples are in the form of predicates applied to constant terms, as in logic-based knowledge bases. 
Their goal is then to learn, in a \textbf{supervised} manner, ABA 
frameworks. This \tags{segmented} approach amounts to 
``identifying rules, assumptions and their contraries (adding to those in the given background knowledge) that cover argumentatively the positive examples and none of the negative examples according to some chosen semantics''.
The works focus on theoretical advancements of learning and transforming logic-based rules that later give rise to \tags{ABA} frameworks for argumentative reasoning. 
In this sense, the approach can be said to use \tags{inductive ML} for automated \tags{generation} of argumentation frameworks. 
However, no experimental evaluation on ML-targeted datasets or comparison with inductive ML (such as inductive logic programming) systems has yet been carried out. 

We next take a look at a particular work that uses an NN architecture to learn argumentation frameworks, this time of the abstract kind.

The argumentative approach to review aggregation introduced by \cite{Cocarascu_19} is supported by ML in a \textbf{segmented} manner. 
In this work, \textbf{QBAFs} are used to represent critics' reviews on movies, i.e.\ \tags{textual data}, in order to provide an explainable method for their aggregation.
An ontology is used to provide a skeleton for arguments in the QBAF, before the arguments, attacks, supports and base scores are then extracted from their reviews. \tags{Supervised} learning via \tags{long-short term memory} NNs is used to extract votes on arguments from the critics, which allows the calculation of the base score (intrinsic strength) on arguments. 
Also, \textbf{topic modeling} is used to \tags{generate} some of the arguments in the QBAF from the reviews.
The QBAFs, evaluated by means of a gradual semantics, then provide users with an aggregation of the reviews supported by explanations comprising components of the QBAF.
The QBAFs and the arguments' evaluations are shown theoretically to satisfy desirable argumentative properties and experimentally to be comparable to the aggregations popularised by the movie review website \emph{Rotten Tomatoes}.

Finally, one recent work looks at the enforcement problem, i.e.\ how to modify an AF to enforce the specified argument acceptability, from the point of view of learning AF modifications from generated AFs, rather than solving enforcement analytically.

Concretely, \cite{Craandijk_22} use deep \tags{RL} to learn which attack relations between arguments should be added or deleted in order to enforce the acceptability of (a set of) arguments.
In particular, the authors present an Enforcement GNN (EGNN), an \tags{approximate} approach, that efficiently learns solutions, i.e.\ modifications to \tags{AFs} that enforce argument (non-)acceptance, which only need to be verified for correctness by existing argument acceptability algorithms. 
In more detail, the authors formulate the problem of learning \tags{enforcement} as a \tags{MDP} where a state represents a modification of a given AF, the actions amount to adding or deleting attacks, with the obvious transition function taking an AF to its modification, and the only non-negative reward is given at the AF with the correct arguments enforced. 
To learn approximately optimal policies of enforcement in such a vast state and action space, the authors employ deep \tags{Q-learning}, whereby the Q-values of states are predicted by a \tags{GNN}. 
The (fully-connected) GNN in question vectorially represents arguments that should be enforced (as nodes) and existing attack relations (as edges) and uses message passing to iteratively update node vectors. 
They compare their EGNN to two other relevant deep learning models as well as to symbolic solvers by training the ML models on \tags{AF data} generated by the ICCMA generators and testing on random sets of argument to be enforced on randomly generated AFs. 
The empirical evaluation shows that the proposed EGNN model beats the other deep learning models in terms of generalisation ability (i.e.\ correctly learning enforcements) and essentially outperforms symbolic solvers in terms of efficiency (solutions found within a time limit) while being near optimal (steps taken to find a solution).

While the above works focus on generating argument graphs, the next set of works use ML to support computation in given argument graphs.

Probably the first work to use ML to support \tags{computation} in argumentation is that of \cite{Garcez_05}. 
There the authors establish an \tags{approximate} mapping from \tags{VAFs} to \tags{NNs} with a single hidden layer operating under \tags{supervised} learning with \tags{tabular data}, showing a correspondence between the two. 
The authors first introduce a Neural Argumentation Algorithm, which, given a VAF, produces a corresponding NN, the uniqueness of which is guaranteed.
The authors show a number of theoretical results, first demonstrating the fact that the algorithm may not be able to compute the NNs corresponding to cyclic VAFs.
In order to resolve this, the authors define a form of learning based on enforcement, i.e.\ a method for implementing changes such that a given state of acceptance or rejection for the arguments is achieved. 
This learning uses backpropagation to indicate changes which need to be made, i.e.\ ``arguments learnt'', in order for specified arguments to be accepted, and the approach is thus shown to model cumulative (accrual) argumentation over time.

More recent works explore how to use GNNs for computing acceptability of arguments in AFs, typically without correctness guarantees, but possibly with high accuracy and speed.

\cite{Kuhlmann_19} use convolutional \tags{GNNs} (GCNs) on \tags{AFs} to \tags{compute} whether an argument is included in the preferred semantics, thus proposing a method to \tags{approximate} reasoning with abstract argumentation. While the original GCN used was designed for semi-supervised training, in this work all nodes in the \tags{AF data} are labelled, thus the training is \tags{supervised}. Further, the number of incoming and outgoing attacks per argument are also used as features. The authors report moderate results on data from ICCMA 2017 and provide several factors that impact the performance (e.g.\ size of the training dataset, class imbalance, as well as hyperparameters).

Another work employing \tags{GCNs} in an \tags{approximated} fashion for \tags{computation} in argumentation is that of \cite{Malmqvist_20}. The GCN is used to determine sceptical and credulous acceptability in \tags{AFs}. 
The \tags{supervised} training involves \tags{AF data}, with each AF representing a single connected component in the graph, being fed to the GCN. The authors use randomised training, forcing the GCN to learn to generalise based on the structural properties of the graphs, and test their method on data from ICCMA 2017.
They are able to improve results using the randomised training approach, however the disparities between the positive (accepted) and negative (rejected) acceptability remain unsolved.

The approach of \cite{Craandijk_20} also uses ML with \tags{supervised learning} in an \tags{approximated} manner, where the authors propose an Argumentation \tags{GNN} (AGNN) to \tags{compute} the (likelihood of the) acceptance of arguments in an \tags{AF}. 
In particular, the authors consider the  credulous and sceptical acceptance of arguments under the grounded, preferred, stable, and complete semantics. They experiment with \tags{AF data} generated using ICCMA 2020 generators and show that AGNNs can almost perfectly predict the acceptability under different semantics and their method is able to scale for larger AFs. Further, the proposed AGNNs can be used to guide a basic search for extensions.

Interestingly, \cite{Kuhlmann_22} follow-up on the work of \cite{Craandijk_20} and suggest some sceptical conclusions. 
The authors there study the problem of data selection for generating and computing AFs, particularly the problem of sceptical acceptance under preferred semantics.
They test the AGNN of \cite{Craandijk_20} on an alternative training set to achieve better performance results, and additionally conduct the same tests on an NN model that is a part of the AGNN. 
The results suggest that the AGNN does not learn to solve the problem of sceptical acceptance, but instead learns specific properties of the benchmarks that allow the model to perform relatively well. The authors conclude the same about the NN model also.
Further, they perform experiments with simpler ML models, specifically 5-Nearest Neighbours, NBCs, DTs and RFs, that mainly use simple graph-theoretic features of AFs, and conclude that those models perform rather well on the AF computation task. The authors suggest that the more complex models such as the AGNN tend to simply learn and make use of those features.

We finally see a proposal that retains the argumentative machinery but suggests to use ML to predict which argumentation solver would be best to compute extensions for a given framework.

\cite{Klein_22} perform an experimental study where \tags{supervised} ML is used in an \textbf{approximate} 
manner to select the fastest algorithm to compute skeptical acceptance under preferred semantics in \textbf{AFs}. In particular, the evaluation setup involves several classic ML models (\tags{k-Nearest Neighbours, NBCs, RFs, and SVMs}) and modern \tags{GNNs} (Graph Convolutional Networks, Graph Isomorphism Networks, and GraphSage) for predicting the fastest sound and complete argumentation solver (one out of three). 
The classic ML classifiers used the number of vertices, density and the minimum degree value of the directed graph as well as in- and out-degrees of the query arguments as features, whereas the GNN models were off-the-shelf (not using any precomputed graph, node or edge features); the labels were assigned to AFs according to which of the three solvers is actually the fastest to solve the problem of sceptical acceptance under preferred semantics. 
The authors experimented on \tags{AF data} with 6200 AFs randomly generated with ICCMA 2017 generators, showing the GNN models slightly outperform the classic ML ones, but are typically slower.

\subsubsection{For Analysing or Replacing Argumentation}
\label{sssec:combining}

\begin{table}[H]
\scriptsize
    \centering
    \begin{tabular}{lllllll} \hline
\!\!\!\!\textbf{Work} & 
\!\!\!\!\textbf{Type} & 
\textbf{Arg.} & 
\!\!\!\!\textbf{Learn.}\!\!\!\! &
\textbf{ML Model(s)} & 
\textbf{Task} & 
\!\!\!\!\textbf{Data} \\ \hline
\!\!\!\!\citep{Potyka_21} & 
\!\!\!\!Syn & 
QBAF with WE & 
\!\!\!\!S & 
NN & 
An & 
\!\!\!\!Tab \\ 
\!\!\!\!\citep{Potyka_22} & 
\!\!\!\!Syn & 
QBAF with WE & 
\!\!\!\!S & 
NN  & 
Class & 
\!\!\!\!Tab \\
\!\!\!\!\citep{Rosenfeld_16} & 
\!\!\!\!App & 
BAF, Dialogues & 
\!\!\!\!S & 
SVM, DT, NN & 
Rec & 
\!\!\!\!Text \\ 
\!\!\!\!\citep{Donadello_22} & 
\!\!\!\!Seg & 
Informal, Dialogues & 
\!\!\!\!S & 
SVR, RF & 
Rec & 
\!\!\!\!Text \\ 
\hline
    \end{tabular}
   \caption{A summary of the ways in which ML has been used for \emph{analysing} or \emph{replacing} argumentation, as detailed in §\ref{sssec:combining}. We show: 
    the type (\underline{syn}ergistic, \underline{seg}mented or \underline{app}roximate); 
    the kind of \underline{arg}umentation used; 
    the type of \underline{learn}ing employed (\underline{s}upervised);
    the ML model(s) used, including support vector regression (SVR); the task undertaken (\underline{an}alysis, \underline{class}ification or argument \underline{rec}ommendation) and the type of data used (\underline{tab}ular or \underline{text}ual).
    }
    \label{tab:combining}
    \label{table:combining}
\end{table}

We next discuss several works that synergistically combine ML and argumentation in a way that one complements the other. These works essentially show correspondence between (feed forward) NNs and QBAFs. In one direction, this allows to use ML for deriving (and computing) new gradual semantics. 
In the other, a QBAF can stand as architecture of an NN and potentially improve learning and/or explainability thereof.

\cite{Potyka_21} shows that \tags{NNs}, trained via \tags{supervised} learning on \tags{tabular data} and in the form of MLPs, can be equivalently seen as \tags{QBAFs with weighted edges}. 
Here, arguments in the QBAFs represent the neurons in the corresponding MLP. 
Then, attacks and supports are drawn between arguments where the connections between the corresponding neurons have weights which are negative or positive, respectively. 
The base score for an argument is then obtained, intuitively, from the corresponding neuron's bias.
A gradual semantics for the QBAF is defined in a way that directly corresponds to the activations which are propagated through the MLP's neurons.
The author then theoretically examines the approach in depth: proving convergence guarantees for cyclic and acyclic cases; suggesting solutions to convergence issues; and assessing the semantics wrt satisfaction of existing properties from the literature. 
In summary, the paper exhibits a \tags{synergystic} approach for general \tags{analysis} whereby ML allows the definition of a novel QBAF semantics and suggests a way of learning QBAFs, and reciprocally, argumentation potentially offers ways of constructing sparse MLPs as well as explaining those which are already trained.

Using the concepts introduced in this approach, \cite{Spieler_22,Potyka_22} have shown how these \tags{QBAFs with weighted edges} with the semantics derived from \tags{NNs}, in the form of MLPs, can be learnt from \tags{tabular data}, again combining argumentation with ML \tags{synergistically}. 
They first restrict the QBAFs to be acyclic, as is the case for the NNs for \tags{classification} via \tags{supervised} learning being somewhat replicated.
The structure of the QBAFs is then learnt using two different meta-heuristics: a genetic algorithm and particle swarm optimisation. 
Finally, the authors train the QBAFs for classification, i.e.\ learning the base scores and edge weights, using backpropagation. 
The activation function can then be seen as embedding the gradual semantics. The authors then undertake an experimental evaluation in classification tasks with the UCI ML Repository. The results show high accuracy for the simple problems, which are similar across the two meta-heuristics, as well as a baseline in the form of a DT method. 
However, the authors argue that using QBAFs is more sparse and easier for a human to interpret, aligning with the proposed reasoning behind the general trend highlighted in §\ref{sssection:explaining}.

The last two works we discuss in this subsection paint the picture where argumentation may be sub-optimal in real-world argumentation scenarios and that one may be better of using off-the-shelf ML models for supporting human-like argumentation.

The first effectively argues that bipolar argumentation is not well-suited to select arguments that people choose in real-world deliberation scenarios and instead suggests using ML for recommending arguments in human discussions. 
Concretely, \cite{Rosenfeld_16} use ML models in a \tags{segmented} manner with \tags{textual data} to predict the arguments selected by humans in real-world discussions, showing superiority of ML models over argumentation theory in the task in terms of both accuracy and user satisfaction. 
Specifically, the authors consider \tags{deliberation dialogues} where participants exchange information and try to reach consensus on some topic. They conduct multiple experiments with human subjects aiming to evaluate whether the participants choose arguments that are justified by \tags{BAFs} that model the deliberations. 
They also define features of arguments in terms of a deliberant's previous use of arguments and argument relevance measures (such as minimum (un)directed paths’ length between arguments) to enable a setting for predicting by means of ML the next argument to choose.
They then study if instead simple \tags{supervised} ML models of \tags{SVMs, DTs} and \tags{NNs} can predict human choices of arguments in deliberations.
In the first experiment, 64 people were presented with a partial deliberation and had to choose one argument out of 4 to continue deliberation. On average, a justified argument (under some classical BAF semantics) was selected only 67.3\% of the time, whereas ML models achieve accuracy above 70\%. 
In the second experiment with 2 corpora and 64 free-form deliberations, 
subjects chose argumentatively justified arguments (likewise, under several classical semantics) no better than chance, whereas ML models reached accuracy similar to that in the first experiment. 
In their third experiment with 72 semi-structured deliberations where 144 participants were restricted to 40 pre-defined arguments to choose from, argumentation again modelled the behaviour no better than random, whereas ML models did significantly better. 
Finally, the authors show in extensive human studies with 204 participants and 102 deliberations how an ML-based model that takes into account argument relevance outperforms argumentation-based models when \textbf{recommending} arguments to put forward next. 
Overall the work shows that bipolar argumentation may host limitations in modelling human deliberations in the real world and that simple ML models of SVMs, DTs and NNs have much better predictive power in deliberations than argumentation, roughly at the level of 75\% and higher, depending on the experiment and setting.

Another work uses \tags{supervised} ML to \tags{recommend} which arguments are
put forward in persuasion settings. 
\cite{Donadello_22} use ML for learning utility functions of the system and user agents involved in a persuasion \tags{dialogue} with \tags{informal argumentation} in an automated persuasion system. 
In a state-of-the-art persuasion setting that uses utility maximisation across arguments of both the system (persuader) and the user (persuadee), the utility functions are typically manually specified by experts. 
Instead, the authors' goal is to be able to learn those utility functions from known utilities. 
To this end they use \tags{support vector regression} to estimate utility of arguments for a new user given those for other users. 
Where further utility elicitation is done via questionnaires, the authors aim to minimise the number of questions posed to the user. 
To do so, they cluster users with similar utility functions and employ \tags{RFs} to identify the best questions for regressive utility estimation. 
It is overall a \tags{segmented} method where ML outputs as utility functions are used to select the most persuasive arguments. The authors evaluate it on synthetic, \tags{textual data} using both abstract and realistic (healthy diet) simulated persuasion dialogues. The experiments show the approach to be promising for learning utilities when data is limited, for the purpose of improving automated persuasion systems. 

\subsubsection{Discussion}

\textit{The use of inductive ML to generate argumentation frameworks shows much promise, but is still to be proven with real datasets.}
The two approaches for generating SA frameworks from tabular data using inductive ML give novel approaches to automated reasoning which could serve as the basis for fruitful directions of future research, given their interesting theoretical results.
If the next steps of proving that this theory translates to real-world datasets at scale are made, this could provide a significant advancement for neuro-symbolic systems.
In the case of learning arguments and relationships from text, there seem to be more issues, e.g.\ due to the need to engineer features and the free-form nature of textual data. However, in settings where the semantics of potential arguments drawn from text are relatively constrained, e.g.\ in review aggregation in a specified domain, there seems to be the possibility of fully automating these pipelines. This must be undertaken with caution, however, as in some domains it has been shown that ML can effectively replace argumentative systems.

\textit{GNNs' natural fit for approximating argument evaluation has much potential, capitalising on existing AF datasets and generators.}
The use of ML for computing argument graphs, for instance predicting acceptability in AFs, also shows much potential. This is especially the case when analytical solutions are costly but checking the correctness of, for example, argument acceptance, is cheap. Here, relatively fast ML-based prediction can be combined with verification for correctness, falling back to an argumentation solver where needed. 
The works very much benefited from ICCMA datasets and AF generators, an indication that competition for other argumentation frameworks, such as those which are structured, may have very welcome consequences further down the road, if the availability of data (i.e.\ argument graphs) induced creation of hybrid solvers.

\textit{The mapping from QBAFs to MLPs brings benefits in both directions, but looks especially exciting in introducing learnt, explainable classifiers.}
Three works have shown that argumentation frameworks, namely QBAFs, can be used as the architecture of a certain form of ML models, namely NNs, so that argumentation frameworks are learnt from data using standard ML techniques, and the reasoning/inference follows argumentative structure and is more explainable. 
While the theoretical analysis of argumentative properties of MLPs is also very interesting, the capability for learning classifiers which reason argumentatively, maintain high accuracy and are able to explain their outputs faithfully to their inference process would be very exciting for the field of argumentation and beyond.

\section{Conclusions}
\label{sec:conclusions}

This chapter overviews a growing body of research on cross-fertilisations between argumentation and ML. 
We have broadly categorised these works into those which can be classed as argumentation for ML, and those which are ML for argumentation. 
The two overarching objectives in the former set of works are either improving or explaining ML models' inference, though the two are not mutually exclusive and most of the works tackle the two in tandem. 
The second set of works are less homogeneous, with objectives ranging from using ML to generate argumentation frameworks, to approximating semantics or recommending arguments to users without the need for an argumentation semantics. 
We have shown that this integration of symbolic and data-driven approaches remains a highly active yet open-ended area of research, with great potential along numerous avenues. 
However, challenges still remain for argumentative techniques to demonstrate definitive advantages in the realm of ML. Progress is needed along multiple dimensions, notably regarding scalability and bridging the gap with real human argumentation via user studies. 
With ML capabilities rapidly progressing, it is clear that there is still a role to be played in the advancement of AI by argumentation, given its prowess in explainability and commonsense reasoning, two factors which are notably lacking in even state-of-the-art ML models.

\bibliography{references}

\end{document}